%% file: acl_latex.tex
\documentclass[11pt]{article}

\usepackage[final]{acl}

\usepackage{times}
\usepackage{latexsym}
\usepackage{amsmath}
\usepackage{multirow}
\usepackage{listings} 
\usepackage{xcolor} 

\usepackage[T1]{fontenc}

\usepackage[utf8]{inputenc}

\usepackage{microtype}

\usepackage{inconsolata}

\usepackage{graphicx}

\usepackage{multirow}
\usepackage{booktabs}
\usepackage{graphicx}
\usepackage{subcaption}
\usepackage{fvextra}

\lstdefinestyle{prompt}{
    basicstyle=\footnotesize\ttfamily,
    breaklines=true, 
    breakindent=1em,
    columns=fullflexible,
    keepspaces=true,
    frame=single, 
    framesep=4pt,
    xleftmargin=3pt,
    xrightmargin=3pt,
    aboveskip=4pt,
    belowskip=4pt,
}

\lstdefinestyle{pycode}{
    language=Python,
    basicstyle=\footnotesize\ttfamily, 
    keywordstyle=\bfseries,
    commentstyle=\itshape,
    showstringspaces=false, 
    keepspaces=true,
    columns=fullflexible,
    breaklines=true,
    breakatwhitespace=true,
    tabsize=4,
    frame=single,
    framesep=5pt,
    xleftmargin=5pt,
    xrightmargin=5pt,
    aboveskip=6pt,
    belowskip=4pt,
}
  
\title{Perception, Layout, and Validation: Calibrated Confidence for Reliable Straight-Through Processing of Financial Documents}

\author{
 \textbf{Yichao Jin},
 \textbf{Yushuo Wang}, 
 \textbf{Yuxuan Han},
 \textbf{Kwan Ching Yee Sonia},
 \textbf{Weiyang Song}, \\
 \textbf{Chiu Jin-Chun Kent},
 \textbf{Wong Chong Hwee},
 \textbf{Wong Tiong Kiat},
 \textbf{Kenneth Zhu Ke},
 \textbf{Jingyuan Zhao}
\\
\texttt{\{JinYichao, YushuoWang, YuxuanHan, SoniaKwan, WeiyangSong,}\\ 
\texttt{KentChiu, ChongHweeWong, WongTK, KennethZhu, JingyuanZhao\}@ocbc.com}
\\
 \texttt{OCBC, Singapore}
}

\begin{document}
\maketitle
\begin{abstract}
Straight-through processing (STP) on extracted key-value fields from financial documents without human review requires a calibrated probability together with a bounded guarantee on the residual error of the auto-approved tier. The emergence of modern Vision Language Models (VLMs) provides an out-of-the-box capability for extracting the key-values, but their verbalized confidence signals are unreliable and weakly track field correctness.

This paper introduces a decomposed confidence layer along three interpretable channels, including perception, layout, and validation. Together with a final conformal risk control, the score can be used for reliable STP of financial documents. The method is validated on three public datasets covering real invoices, synthetic invoices, and ad-buy forms, using two different VLM families (Qwen3.6-27B and Gemini-3.1-Flash-Lite). Our decomposed score consistently improves the separation of correct from incorrect extractions, substantially raising the AUROC from 0.54–0.74 for VLM verbalized signals to 0.90-0.99 with contributions from all three designed channels. Crucially for industrial deployment, this enables usable STP. The native VLM confidence signals could clear only 0.1\%-7.0\% of fields under risk control at a target error of $\leq 10\%$. In contrast, the proposed method auto-approves 49-72\% of fields while holding the empirical error of the accepted tier at or below the target.

\end{abstract}

\input{sections/intro}
\input{sections/related_work}
\input{sections/method}

\input{sections/experiment}
\input{sections/conclusion}



\bibliography{custom}

\appendix
\input{sections/appendix}

\end{document}

%% file: sections/intro.tex
\section{Introduction}

Financial back-office operations, such as accounts payable and client onboarding, have to process large volumes of semi-structured documents (e.g., invoices and trade orders) under tight time constraints. Currently, in most financial institutions, manual verification of the extracted fields still dominates the per-document cost. Therefore, \emph{Straight-through processing} (STP), which auto-approves the extractions with minimal human review, becomes the key driver of productivity and scalability.   

In a financial setting, STP carries a stringent bar. In particular, a single wrongly auto-approved value (e.g., a transposed amount or a misread account number) is a critical financial and compliance error. Safe STP thus requires two things beyond a point extraction. First, a per-field probability that is \emph{calibrated}. Second, a \emph{distribution-free bound} on the residual error of the auto-approved tier. Together, these hold the error rate among accepted fields below an operator-chosen budget.

Modern vision-language models (VLMs) have largely solved the extraction step. Prompted with a page image and a field schema, they emit values directly and usually accurately, without task-specific training. What they lack is a reliable confidence signal. A VLM \emph{can} verbalize a confidence score, but these are often saturated and over-confident. On our data, verbalized confidence separates correct from incorrect extractions with an AUROC of only $0.54$–$0.74$ (i.e., at times worse than chance). Token log-probabilities, where exposed, sit over generated sub-word tokens and track field correctness only weakly. Even sampling-based self-consistency \cite{wang2022self} is blind to \emph{systematic} errors, where the model returns the same wrong value on every sample. Consequently, a VLM-only STP system is caught between auto-approving almost nothing and auto-approving errors.

We introduce a decomposed confidence layer for VLM key-value extraction, which scores each extracted field along three interpretable channels. The \textbf{Perception} channel checks if the evidence is legible from region-local image quality, VLM-OCR agreement, and value-string cues. The \textbf{Layout} channel asks if the value sits where this field normally appears. We retrieve historical layouts similar to the current page, and score the field's anchor-relative position under a retrieval-conditioned spatial prior. A value in an unusual location would be flagged even when read perfectly. Lastly, the \textbf{Validation} channel applies type, checksum, and rules, and checks cross-model agreement against a second independent VLM. A gradient-boosted model fuses the three channels into a single probability, with per-field reason codes. We wrap this score in a Learn-Then-Test conformal procedure \cite{angelopoulos2025learn}, which certifies a distribution-free bound on the error rate of the auto-approved tier. Finally, when two VLMs disagree, we select the value our layer scores higher. This resolves the disagreement rather than rejecting it, raising STP throughput at the same guaranteed error.

We validate the method on three public datasets. They span real invoices from DocILE \cite{vsimsa2023docile}, synthetic invoices from FATURA \cite{limam2023fatura}, and advertising order forms from VRDU \cite{wang2023vrdu}, using two VLM families, Qwen3.6-27B \cite{qwen36_27b} and Gemini-3.1-Flash-Lite \cite{gemini31flashlite}. Our main contributions include 

\begin{itemize}
   \item A decomposed and interpretable confidence layer consisting of perception, layout, and validation channels. The fused conformal score reliably reflects the field correctness with a distribution-free error guarantee. Experiments show a substantial AUROC lift from 0.54–0.74 (VLM baseline) to 0.90--0.99.
   
   \item A cross-model verification-and-selection mechanism, that resolves disagreements between two VLMs, further lifts the guaranteed STP coverage on top of fusion of the three channels. The full layer auto-approves 49-72\% of fields at a $\leq 10\%$ target error, versus 0.1-7.0\% from native signals.
   
   \item Per-channel and per-feature SHAP attributions that make each accept and reject decision auditable, as required for model-risk and governance review in a regulated setting.
 \end{itemize}

%% file: sections/related_work.tex
\section{Related Work}

\paragraph{Document key-value extraction.}
Early field extractors coupled OCR with layout-aware encoders, such as the
LayoutLM family \cite{xu2020layoutlm, xu2021layoutlmv2, huang2022layoutlmv3} and OCR-free models like Donut \cite{kim2022ocr}. These require task-specific fine-tuning and predict grounded spans. Modern VLMs instead read a page image and emit field values directly, without training. However, the extracted values are often ungrounded without reliable boxes and confidence scores. 

Our proposed layer is agnostic to the extractor, scoring the confidence of already-extracted values and recovering grounding from OCR spans.

\paragraph{Confidence and calibration for extraction.}
Confidence for extraction and generation has relied on token log-probabilities, verbalized confidence \cite{xiong2024can}, and sampling-based self-consistency \cite{wang2022self}. For VLM field extraction these signals are weak. The closest line fuses heterogeneous signals (i.e., log-probabilities, OCR quality, image quality, and coarse spatial features) into a single learned score and reports an aggregate AUROC \cite{kumar2026beyond}. 

Our work differs in three ways. First, we add a retrieval-conditioned layout prior as a first-class signal, rather than coarse spatial features. Second, we keep the score interpretable through a per-channel decomposition with reason codes. Third, we target a distribution-free straight-through-processing guarantee, not only a ranking metric.

\paragraph{Conformal risk control and selective prediction.}
Selective classification abstains on low-confidence inputs \cite{geifman2017selective}. Split conformal prediction turns any score into distribution-free coverage \cite{vovk2005algorithmic}. Learn-Then-Test \cite{angelopoulos2025learn} and conformal risk control \cite{angelopoulos2024theoretical} extend this to control general risks, including the selective error of an accepted set through multiple-hypothesis testing. 

We use Learn-Then-Test to certify a distribution-free bound on the error rate of the auto-approved tier, class-conditional by field type. Distinctively, our retrieval score is both an input to the confidence model and the variable that indexes the calibration regime, linking template familiarity to the risk-control guarantee.

%% file: sections/method.tex
\section{Problem Statement and Methodology}
\label{sec:statement}

\begin{figure*}[t]
  \centering
  \includegraphics[width=\textwidth]{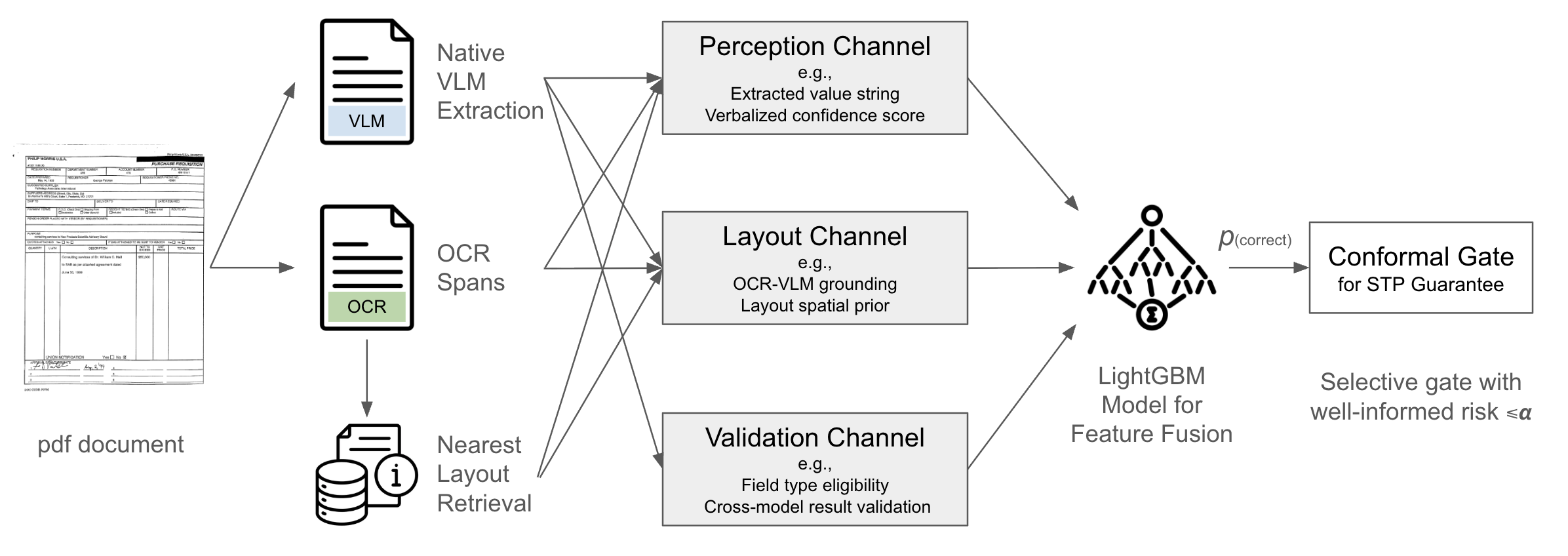}
  \caption{End-to-end confidence layer. The VLM extracts ungrounded field values from the page image. OCR spans recover grounding, and nearest-layout retrieval supplies similar historical templates. Three interpretable channels (i.e., perception, layout, and validation) score each extracted value. A gradient-boosted model (LightGBM) fuses them into a calibrated probability $p(\mathrm{correct})$, which a conformal selective gate turns into an auto-approval decision at a guaranteed selective risk $\le \alpha$.}
  \label{fig:pipeline}
\end{figure*}

Figure \ref{fig:pipeline} shows the end-to-end confidence layer. The input is a pdf document, where a VLM extracts field values directly with verbalized confidence score. In parallel, OCR yields word spans, and a retrieval step fetches historical layouts similar to the page. Then three channels examine the corresponding features to score each extracted value. In particular, the \textbf{perception} channel checks if the evidence is legible. The \textbf{layout} channel checks whether the value sits where the field normally appears. The \textbf{validation} channel applies content rules and cross-model agreement. Subsequently, a LightGBM model fuses the three channels into a single probability $p_f$. Finally, a conformal selective gate turns $p_f$ into an auto-approval decision with a controlled error rate. We detail each component below.

\subsection{Grounding recovery}
A VLM reads a document image and a field schema, and returns a value $v_f$ and its verbalized confidence score $s_f$ for each field type $f$. Our layer assigns each extracted field $(f, v_f, s_f)$ a calibrated probability $p_f$ that the value is correct, using the page and its recovered OCR grounding.

We first recover grounding from OCR word spans, where each span is a candidate $c$ with printed text $\mathrm{text}(c)$ and a bounding box. The candidate set for field $f$ collects the spans whose canonical text matches the extracted value
\[
  C_f = \{\, c : \mathrm{canon}_f(\mathrm{text}(c)) = \mathrm{canon}_f(v_f) \,\}
\]
where $v_f$ is the VLM value and $\mathrm{canon}_f$ is a per-type canonicalizer which handles numeric tolerance, ISO-8601 dates, and normalized strings. $|C_f|{=}0$ means the value is not on the page (a strong error signal) and $|C_f|{>}1$ means the value is ambiguous. The details of canonicalizer design are illustrated in Appendix \ref{app:canon}.

\subsection{Perception channel}
The perception channel asks whether the evidence is legible and read correctly with 11 features. From the value string we take length, token count, digit ratio, and confusable-glyph mass (e.g., $0/O$, $1/l$, $5/S$). At the candidate box, we take region-local image quality (i.e., Laplacian variance, contrast, ink density, and glyph height) and the edit distance between $v_f$ and the OCR text there. We also include the VLM's verbalized confidence. 

\subsection{Layout channel}

\begin{figure*}[t]
\centering
\includegraphics[width=\textwidth]{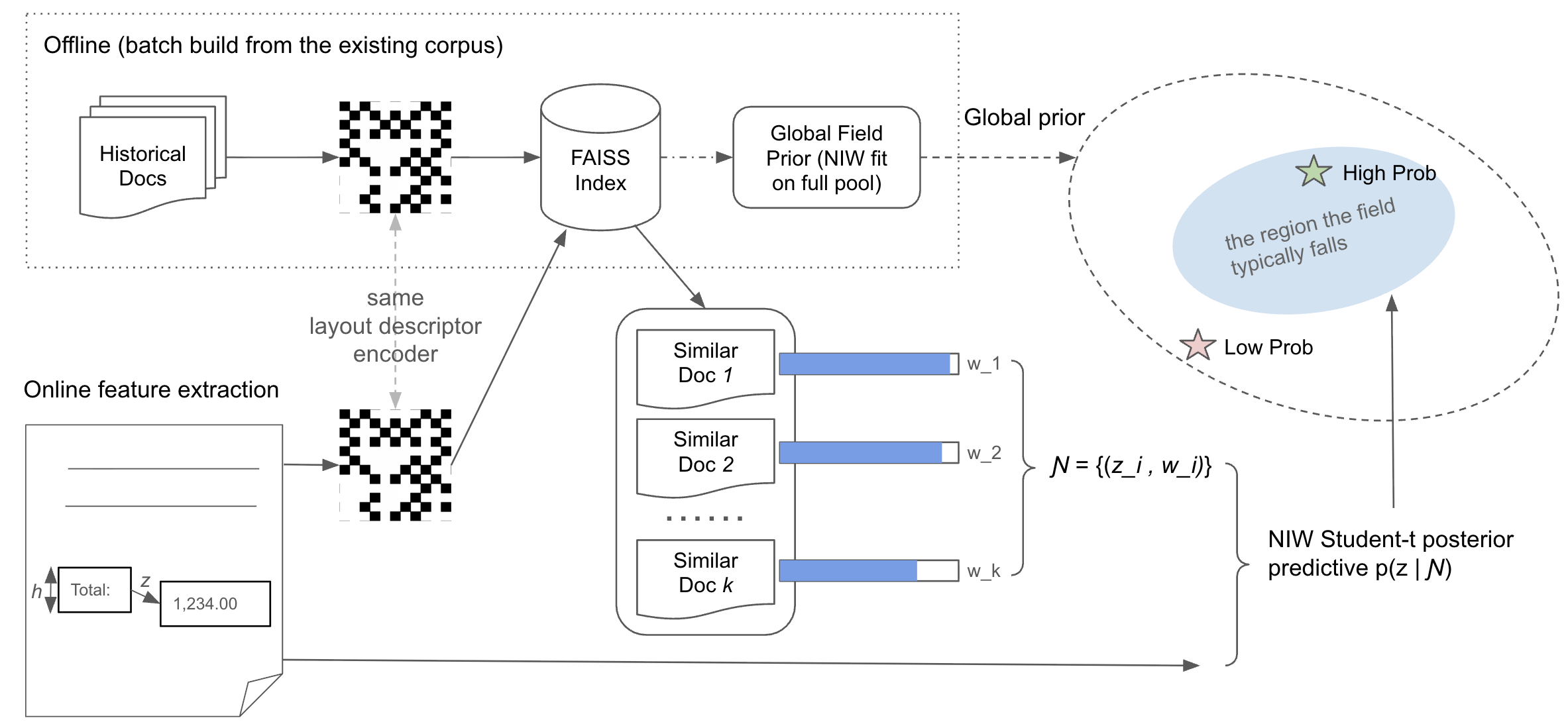}
\caption{Layout-channel drill-down. \textbf{Offline} (top): each historical document
is encoded by the layout-descriptor encoder $\phi$ into a FAISS index, and the field positions across the corpus are pooled to fit a global Normal--Inverse--Wishart (NIW) prior. \textbf{Online} (per field): (1) grounding gives the anchor-relative position $z$ of the value (offset from its label anchor, normalized by anchor height $h$); (2) the query page is encoded by the \emph{same} descriptor $\phi$ and retrieves the $k$ nearest historical layouts, each contributing its field position and a similarity weight, $\mathcal{N}=\{(z_i,w_i)\}$; (3) The global prior updated with the similarity-weighted neighbours $\mathcal{N}$ yields a Student-$t$ posterior predictive $p(z\mid\mathcal{N})$. The shaded blue region is where the field typically lands for this template (tight), while the dashed ellipse is the global prior it falls back to when no neighbours are found (wide). A candidate whose position lies in the blue region scores high, whereas one in the tail (Low Prob) is penalized even when read correctly.}
\label{fig:layout}
\end{figure*}

The layout channel contributes 15 features and is illustrated end to end in Figure~\ref{fig:layout}. It asks whether $v_f$ sits where field $f$ normally appears from documents with similar layout.

\paragraph{Anchor-relative geometry.}
We encode a candidate's position relative to a nearby label anchor rather than in absolute page coordinates. The anchor is chosen geometrically, not semantically. Among all OCR words on the page (excluding the value's own tokens), we take the nearest label-like word under a fixed preference order. A word on the same text line and to the left of the value (within $0.6h_v$ vertically, where $h_v$ is the value height) is preferred, scored by its horizontal distance. So the closest same-row-left word wins. This captures the common ``\texttt{Label: value}'' layout. If not found, we fall back to a word directly above and roughly column-aligned (within $3h_v$ horizontally). This handles column-header layouts. The matching uses no keyword list or field-type dictionary, so it is robust to unseen templates. Given the anchor at center $center(a)$ with height $h_a$, the position feature is the offset normalized by anchor height as
\[
  z = \frac{\mathrm{center}(v) - \mathrm{center}(a)}{h_a}
\]
$z$ is invariant to translation and to scale across templates and page sizes. When no anchor qualifies, $z$ falls back to the offset from the page centre.

\noindent\textbf{Retrieval.} We encode each page as a binary token-mask descriptor (details in Appendix~\ref{app:descriptor}) and index the historical documents with FAISS~\cite{douze2025faiss}. To inference a new page we retrieve the $k$ nearest layouts with similarity weights $\{w_i\}_{i=1}^{k}$. We summarize familiarity by the top similarity $s_{\mathrm{match}} = \max_i w_i$ and a soft effective neighbour count $k_{\mathrm{eff}} = \big(\sum_i w_i\big)^2 / \sum_i w_i^2$.

\noindent\textbf{Retrieval-conditioned spatial prior.} We model $z$ for each field type with a Normal--Inverse--Wishart (NIW) \cite{murphy2007conjugate} prior on its mean and covariance, fit on the global field-conditional pool. We update the prior with the retrieved neighbours' positions, weighted by similarity, and marginalize to a Student-$t$ posterior predictive $p(z \mid \mathcal{N})$. A single shrinkage hyperparameter $\kappa_0$ governs how fast neighbours override the global prior. The behaviour is self-regularizing in familiarity. In particular, at $k_{\mathrm{eff}}{=}0$ the posterior collapses to the global field prior (cold start); small $k_{\mathrm{eff}}$ yields heavy tails and appropriately low confidence; large $k_{\mathrm{eff}}$ concentrates around the template-specific location.

\noindent\textbf{Non-circular scoring.} A value can appear at several places on the page, so $C_f$ may contain multiple positions $z_c$. We score the field's placement with the layout score $s_L$, the log-average posterior density over these occurrences as
\[
  s_L = \log \frac{1}{|C_f|}\sum_{c\in C_f} p(z_c \mid \mathcal{N}).
\]
Averaging rather than taking the maximum avoids a selection bias. Selecting the maximum alone would make a value look correct merely because one of its occurrences happens to land in a high-density region. Instead, the spread across occurrences is surfaced separately as ambiguity signals, derived from the occurrence distribution $\pi_c \propto p(z_c\mid\mathcal{N})$ and its entropy $H_f$. We expose three instances of $s_L$ into the feature set, including the retrieval-conditioned posterior \texttt{s\_l\_marg}, the global prior with retrieval off \texttt{s\_l\_cold} whose gap to the primary exposes the retrieval lift, and under an absolute-position prior \texttt{s\_l\_abs} as a weak baseline.

\begin{table*}[t]
\centering
\small
\setlength{\tabcolsep}{6pt}
\begin{tabular}{@{}lcccccc@{}}
\toprule
& \multicolumn{2}{c}{DocILE} & \multicolumn{2}{c}{FATURA} & \multicolumn{2}{c}{VRDU}\\
\cmidrule(lr){2-3}\cmidrule(lr){4-5}\cmidrule(lr){6-7}
Method & Qwen & Gemini & Qwen & Gemini & Qwen & Gemini\\
\midrule
Self-consistency        & 0.728 & 0.660 & 0.605 & 0.544 & 0.629 & 0.560\\
VLM baseline            & 0.736 & 0.702 & 0.614 & 0.597 & 0.597 & 0.550\\
\;\textbf{+ P/L/V fusion}        & 0.911 & 0.908 & 0.990 & 0.996 & 0.921 & 0.934\\
\;\textbf{+ cross-model}         & \multicolumn{2}{c}{\textbf{0.915}} & \multicolumn{2}{c}{\textbf{0.997}} & \multicolumn{2}{c}{\textbf{0.928}}\\
\bottomrule
\end{tabular}
\caption{AUROC for predicting extraction correctness across four approaches, on three datasets and two VLM families. The cross-model row fuses both VLMs into one output (one value per dataset).}
\label{tab:auroc}
\end{table*}

\subsection{Validation channel}
The validation channel applies content rules and contributes 8 features. It checks type and format validity, ISO-4217 currency and date
validity, checksums where a field admits one (e.g.\ IBAN mod-97, ISIN), and
arithmetic residuals across amount fields (e.g., net $+$ tax $=$ gross). It also records cross-model agreement on whether a second, independent VLM returns the same canonical value. 

\subsection{Feature fusion model}
A gradient-boosted model fuses the three channels into a single probability $p_f$. We use LightGBM over the concatenated per-channel features, learning correlated interactions. The model is tabular and yields SHAP attributions, so each accept/reject decision carries per-feature reason codes for informative model-risk review. The full feature list can be found in Appendix \ref{app:features}.

\subsection{Conformal selective gate}
A calibrated probability is not yet a decision. The gate accepts a field iff $p_f \ge \lambda$, where $\lambda$ is chosen on a held-out calibration set so that the \emph{selective risk} (error rate among accepted fields) stays at or below an operator-chosen budget $\alpha$. Learn-Then-Test \cite{angelopoulos2025learn} makes this threshold search valid, giving a distribution-free guarantee at confidence $1-\delta$. As a result, the auto-approved tier carries error at or below $\alpha$, and everything below the threshold is routed to human review.

%% file: sections/experiment.tex
\section{Experiments}

\begin{figure*}[t]
\centering
\begin{subfigure}[t]{0.32\textwidth}
  \includegraphics[width=\textwidth]{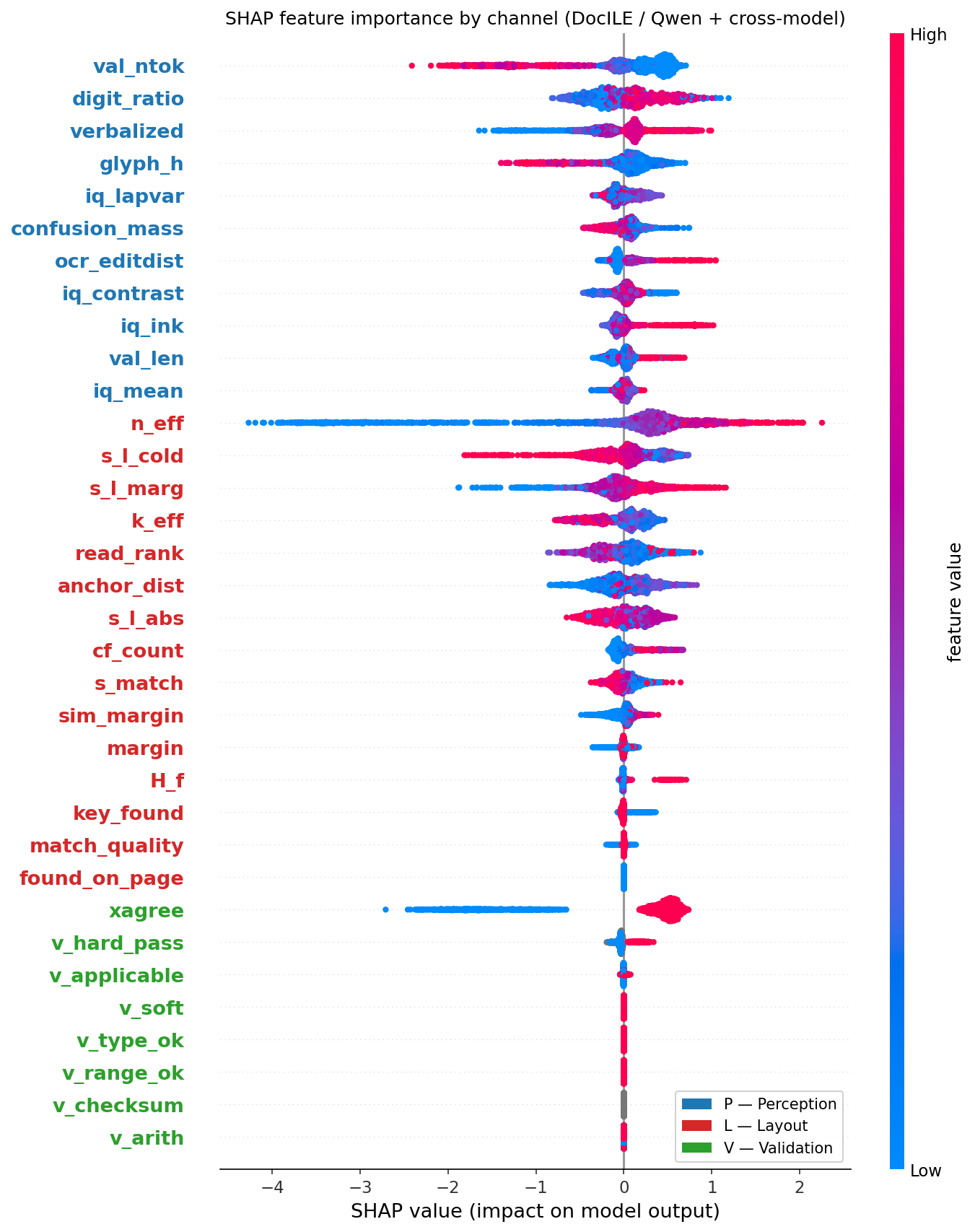}
  \caption{DocILE}\label{fig:shap-docile}
\end{subfigure}\hfill
\begin{subfigure}[t]{0.32\textwidth}
  \includegraphics[width=\textwidth]{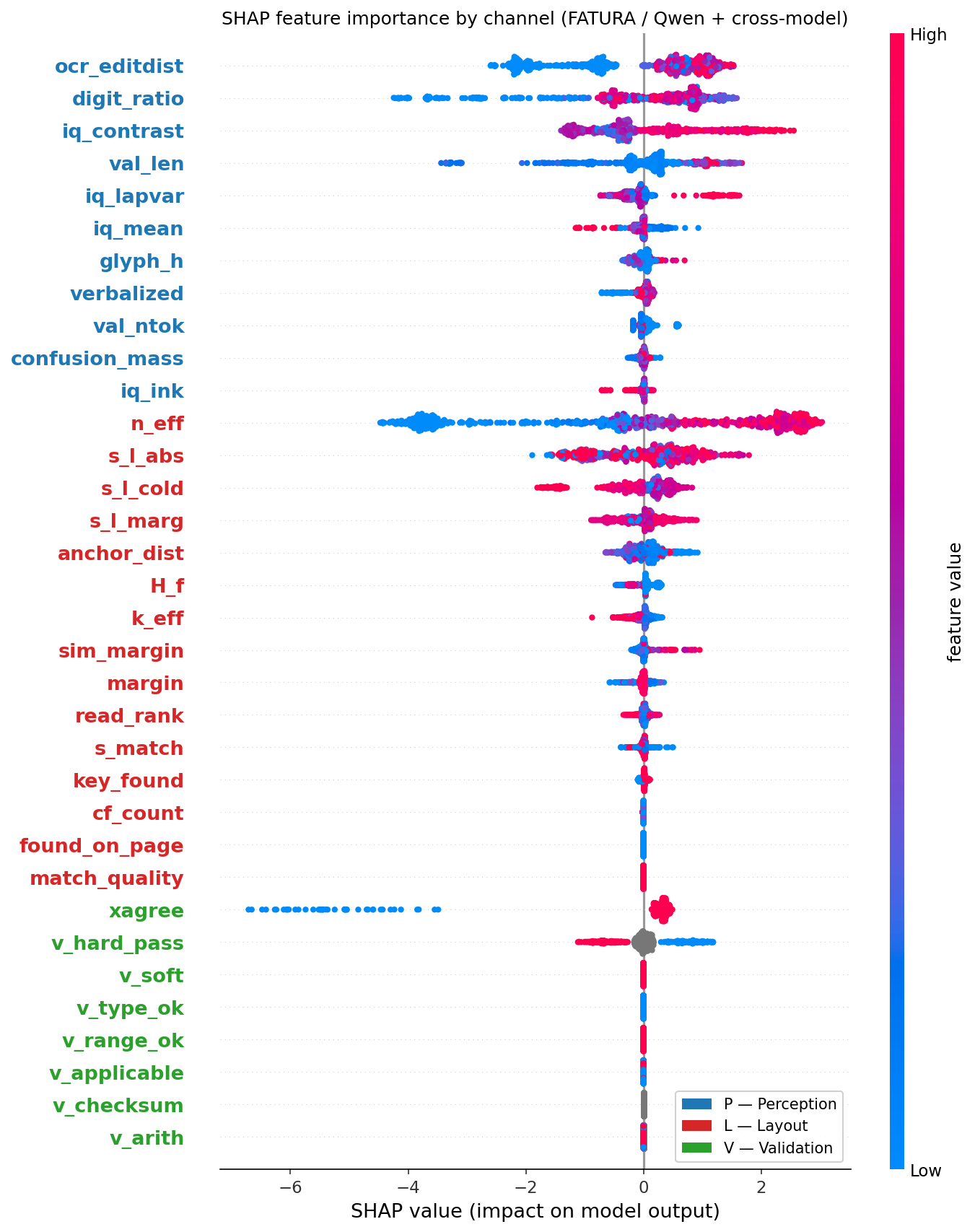}
  \caption{FATURA}\label{fig:shap-fatura}
\end{subfigure}\hfill
\begin{subfigure}[t]{0.32\textwidth}
  \includegraphics[width=\textwidth]{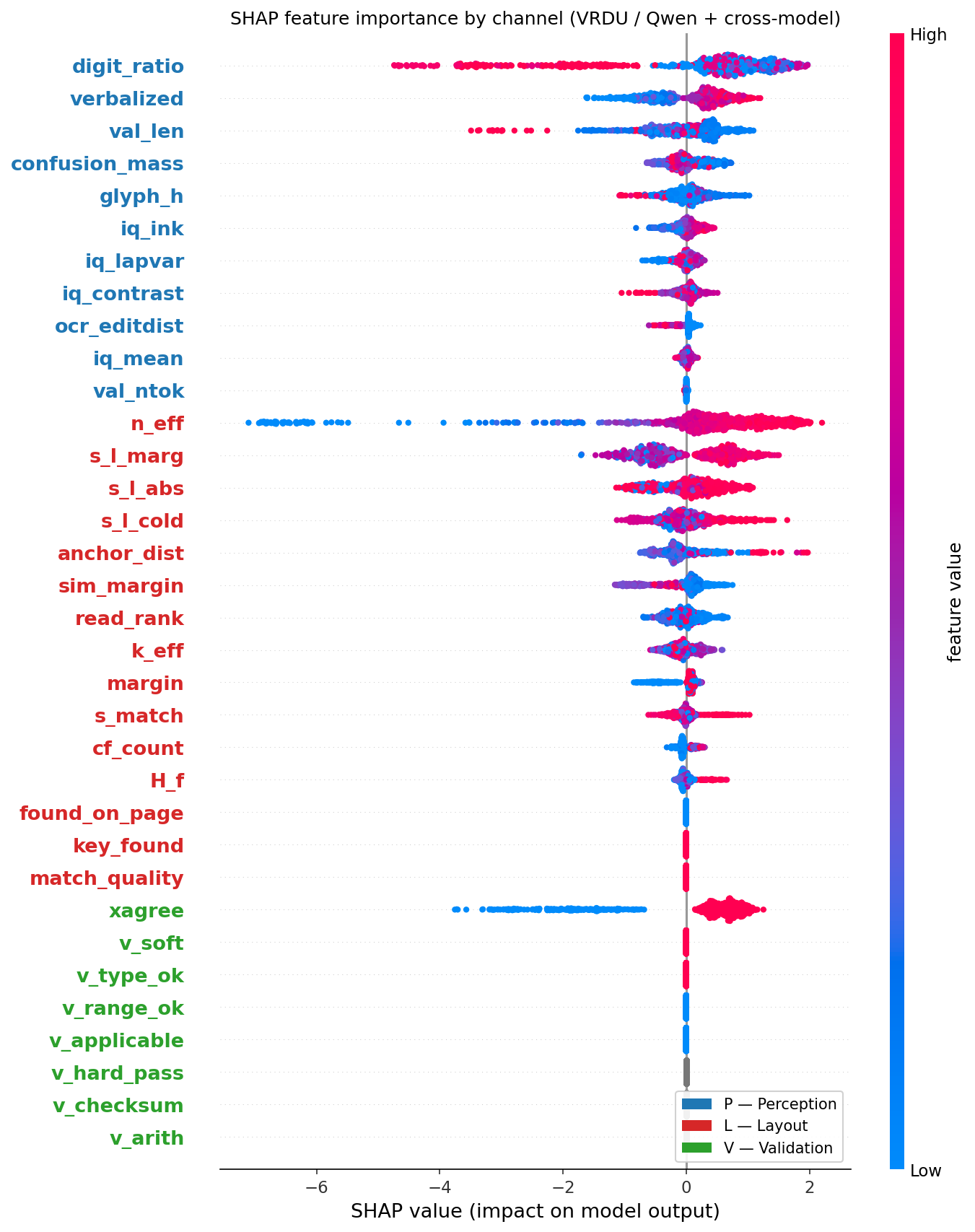}
  \caption{VRDU}\label{fig:shap-vrdu}
\end{subfigure}
\caption{SHAP feature importance by channel (Qwen extractor $+$ cross-model) on the three datasets. Labels are colored by channel (P blue, L red, V green). The channel emphasis shifts slightly with document type: FATURA is perception/OCR-driven, VRDU is layout-driven, DocILE is in between.}
\label{fig:shap}
\end{figure*}

\subsection{Experimental setup}
\label{sec:setup}

\paragraph{Datasets.}
We evaluate three public document collections spanning real invoices DocILE~\cite{vsimsa2023docile}, synthetic invoices FATURA~\cite{limam2023fatura}, and advertising order forms VRDU~\cite{wang2023vrdu}. Each contributes a subset of fields unambiguously under a shared taxonomy (19, 14, and 9 types respectively). The offline layout index is built for each dataset from the training split and is disjoint from the evaluation documents. The details are given in Appendix~\ref{app:datastats}.

\paragraph{VLM extractors.}
To test that the confidence layer is model-agnostic, we run every experiment with two independent VLM families, Qwen3.6-27B and Gemini-3.1-Flash-Lite. A value and its verbalized confidence for each field are returned from a single ungrounded prompt over the page image. We query each document $K_q{=}5$ times to estimate self-consistency
and take the canonical value as the extraction. The full prompts can be found in Appendix~\ref{app:prompt}, and the detailed fusion model configurations can be found in Appendix~\ref{app:fusion}.

\paragraph{Baselines.}
We compare against the confidence signals from the VLM alone (i)
\emph{self-consistency}: the fraction of the $K_q=5$ samples agreeing with the modal value; and (ii) a \emph{VLM baseline}: the native verbalized confidence score. We report our full end-to-end pipeline based on feature engineering from three channels and the fusion layer.

\subsection{Correctness discrimination}
\label{sec:results-auroc}

Table~\ref{tab:auroc} reports AUROC for predicting whether an extracted value is correct. As baselines, the VLM's own confidence signals are weak: self-consistency and the VLM baseline reach only $0.54$--$0.74$ across the six dataset--model settings. Notably, self-consistency is not stronger than the single-call VLM baseline despite costing $K_q$ calls. This is because sampling agreement detects only stochastic errors, where resampling changes the answer, but is blind to systematic ones, where the same wrong value is returned every time. In practice, many VLM extraction errors are systematic, so high agreement reflects determinism, not correctness. 
In contrast, adding the P/L/V channels together with the fusion layer considerably raises AUROC to $0.91$--$0.99$, which holds across all dataset and VLM settings, confirming the pipeline is largely model-agnostic. Moreover, cross-model agreement adds a further increment when both VLMs are available. The lift is largest on FATURA at $0.997$, whose clean synthetic templates make layout and OCR cues highly reliable, and is consistent on the other two real-document collections at $0.91$--$0.93$.

\subsection{What the fusion model learns}
\label{sec:results-shap}

Figure~\ref{fig:shap} shows SHAP attributions for the fused model, grouped and colored by channel. Three patterns hold across all datasets. 
First, the retrieval prior is the strongest signal. The effective-neighbour count \texttt{n\_eff} has the widest impact in every panel, with a long low-value tail that correctly lowers confidence at cold start. And the retrieval-conditioned spatial scores (\texttt{s\_l\_marg}, \texttt{s\_l\_cold}) and anchor distance (\texttt{anchor\_dist}) form a consistent supporting cluster. 
Second, cross-model agreement (\texttt{xagree}) is a clean and high-impact validator that agreement raises confidence and disagreement lowers it.
Third, within perception the value-string and OCR cues do most of the work using token count (\texttt{val\_ntok}), digit ratio (\texttt{digit\_ratio}), glyph height (\texttt{glyph\_h}), and VLM--OCR edit distance (\texttt{ocr\_editdist}). While the VLM's own verbalized confidence ranks only mid-pack, consistent with its weakness as a standalone signal.

Aggregated by channel, layout, and perception dominate ($49\%$ and $39\%$ of overall attribution), with validation contributing the rest. The channel emphasis shifts with document type. FATURA is perception/OCR-driven because of its clean and synthetic templates, VRDU is layout-driven, and DocILE is in between. The top signals are stable across all datasets and VLM families.

The interpretability of the layer is a deployment prerequisite in a regulated financial setting. An automated approval that cannot be explained is, in practice, often not acceptable. Model-risk management and audit functions require that every straight-through decision be traceable to a justifiable cause, where a monolithic neural confidence score is insufficient. Because our fusion model is tabular and its inputs are organized along the interpretable perception, layout, and validation channels, every accept/reject decision decomposes via SHAP into signed per-feature contributions that read as human-legible. 
A field would be routed to review because its value has little retrieval support (e.g., an unfamiliar template), or the two VLMs disagree, or it sits an unusual distance from its label anchor, or the VLM reading diverges from the OCR text. These inspections are exactly the artifacts a model validation or regulator review asks for, allowing a human reviewer to either trust or override the gate on a case-by-case basis. The confidence layer is therefore not merely a score but an \emph{auditable control} in a compliance-bound environment.

\begin{table*}[t]
\centering
\small
\setlength{\tabcolsep}{6pt}
\begin{tabular}{@{}llrrr@{}}
\toprule
Dataset & Method & $\le$5\% err & $\le$10\% err & $\le$20\% err\\
\midrule
\multirow{4}{*}{DocILE}
 & Self-consistency (Qwen/Gemini) & 0.1 / 0.0 & 0.2 / 0.0 & 0.5 / 0.3\\
 & VLM baseline (Qwen/Gemini)     & 0.3 / 0.1 & 1.6 / 0.1 & 12.2 / 0.6\\
 & + P/L/V fusion (Qwen/Gemini)         & 20.7 / 21.4 & 39.3 / 42.6 & 58.0 / 61.1\\
 & \textbf{+ cross-model}          & \textbf{28.1} & \textbf{49.4} & \textbf{66.5}\\
\midrule
\multirow{4}{*}{FATURA}
 & Self-consistency (Qwen/Gemini) & 0.1 / 0.2 & 0.1 / 0.2 & 0.2 / 0.5\\
 & VLM baseline (Qwen/Gemini)     & 0.1 / 1.4 & 0.1 / 7.0 & 0.4 / 18.1\\
 & + P/L/V fusion (Qwen/Gemini)         & 62.4 / 66.6 & 66.3 / 70.5 & 74.8 / 79.5\\
 & \textbf{+ cross-model}          & \textbf{68.4} & \textbf{72.3} & \textbf{81.6}\\
\midrule
\multirow{4}{*}{VRDU}
 & Self-consistency (Qwen/Gemini) & 0.2 / 0.2 & 0.4 / 0.3 & 1.4 / 0.9\\
 & VLM baseline (Qwen/Gemini)     & 1.5 / 0.1 & 2.9 / 0.1 & 10.4 / 0.2\\
 & + P/L/V fusion (Qwen/Gemini)         & 25.7 / 35.9 & 60.5 / 66.9 & 78.7 / 79.7\\
 & \textbf{+ cross-model}          & \textbf{41.3} & \textbf{70.8} & \textbf{83.8}\\
\bottomrule
\end{tabular}
\caption{Straight-through-processing coverage, in terms of the percentage of fields that can be auto-approved while holding the accepted-tier error at or below the target. Native VLM signals clear almost no fields, while the channel fusion and cross-model selection unlock up to 84\% when $\leq 20\%$ error.}
\label{tab:stp}
\end{table*}

\subsection{Straight-through-processing coverage}
\label{sec:results-stp}

Table~\ref{tab:stp} shows the deployment-facing result on the percentage of fields that can be auto-approved while holding the accepted-tier error at or below a target. This is the operative metric in the real regulatory setting where straight-through processing is precision-first. Specifically, a wrongly auto-approved value is a financial or compliance error, whereas a correct value sent to review is only a small efficiency cost. Consequently, the question is not average accuracy but how much can be auto-approved at a guaranteed precision.

In particular, at a $10\%$ error target ($\ge\!90\%$ precision) self-consistency and the VLM baseline auto-approve $0.1$--$7.0\%$ of fields. Whereas the P/L/V fusion auto-approves $39$--$71\%$ at the same target, which is a one to two orders of magnitude increase, driven collaboratively by all three channels (Section~\ref{sec:results-shap}). Moreover, cross-model selection adds a premium tier ($49$--$72\%$, up to $84\%$ at a $20\%$ target), most visibly on VRDU, where resolving disagreements rather than rejecting them lifts coverage from $61$--$67\%$ to $71\%$. The gains hold across both VLM families. Even DocILE, the real and heterogeneous invoices, reaches $49\%$ auto-approval at $\ge\!90\%$ precision from a near-zero baseline. 

Appendix~\ref{app:cases} gives qualitative case studies to make the STP concrete. In each case a VLM returns a wrong value with high verbalized confidence ($\ge\!0.90$), yet our pipeline scores it low and cross-model selection recovers the correct value.

%% file: sections/conclusion.tex
\section{Conclusion}

We presented a decomposed confidence layer for straight-through processing of VLM-extracted fields from financial documents. It scores each field along three interpretable channels (perception, layout, and validation) with a retrieval-conditioned layout prior. A LightGBM model fuses the channels into a calibrated probability with per-field reason codes and a Learn-Then-Test conformal gate turns it into an auto-approval decision with a distribution-free error bound. Across three public datasets and two VLM families, the layer raises correctness-discrimination AUROC from $0.54$-$0.74$ for native VLM confidence to $0.90$--$0.99$, and auto-approves $49$--$72\%$ of fields at a $10\%$ error target where native signals clear almost none. The gains hold across both VLMs showing its model-agnostic. A cross-model selection step resolves disagreements from two VLMs to further lift guaranteed throughput.

The limitations are mild rather than fundamental in target banking settings. First, the layout channel is weakest at cold start. For a genuinely novel template with no similar layouts in the index, the retrieval-conditioned prior collapses to the global field prior and adds little signal. However, in the real operation, the document stream is dominated by recurring templates (e.g., the same forms and counter-parties processed repeatedly). A truly new template is rare, and it degrades gracefully to other channels and, under the conformal gate, is routed to human review rather than mis-approved. Each such document then enters the index, so a new template becomes familiar after only a few occurrences. Second, fitting the fusion model and calibrating the gate require field-level correctness labels. However, financial institutions usually have large volumes of approved historical documents, providing enough supervision. Moreover, mandatory sample-checking of processed items (including auto-approved STP cases) yields a continual stream of audited field-level labels for periodic refresh. Finally, the rule validators add little marginal signal on the reported dataset, where errors are well-formed and no check-summable identifiers appear. Their value should grow on documents such as trade orders containing IBAN / ISIN / BIC.

%% file: sections/appendix.tex
\section{Canonicalization Rules}
\label{app:canon}

\begin{table*}[t]
  \centering
  \footnotesize
  \setlength{\tabcolsep}{5pt}
  \begin{tabular}{@{}lp{\dimexpr\textwidth-2.9cm\relax}@{}}
  \textbf{Feature} & \textbf{Description} \\
  \midrule
  \multicolumn{2}{@{}l}{\textit{Perception Channel --- 11 features}}\\
  \midrule
  \texttt{verbalized}      & \textit{self-report.} VLM verbalized confidence $s_f$, rescaled to $[0,1]$.\\
  \texttt{iq\_lapvar}      & \textit{pixels.} Laplacian variance of the value crop (sharpness; low = blurry).\\
  \texttt{iq\_contrast}    & \textit{pixels.} Pixel contrast of the value crop.\\
  \texttt{iq\_ink}         & \textit{pixels.} Ink density (dark-pixel fraction) of the crop.\\
  \texttt{iq\_mean}        & \textit{pixels.} Mean gray level of the crop.\\
  \texttt{glyph\_h}        & \textit{pixels.} OCR text-box height in pixels (small = tiny/low-resolution print).\\
  \texttt{ocr\_editdist}   & \textit{OCR.} $1-$ string similarity of $v_f$ to the OCR text at the location (VLM--OCR disagreement).\\
  \texttt{val\_len}        & \textit{string.} Character length of the value string.\\
  \texttt{val\_ntok}       & \textit{string.} Whitespace-token count of the value string.\\
  \texttt{confusion\_mass} & \textit{string.} Fraction of chars in the confusable-glyph set ($0/O$, $1/l/I$, $5/S$, $8/B$, \texttt{.}/\texttt{,}).\\
  \texttt{digit\_ratio}    & \textit{string.} Fraction of value characters that are digits.\\
  \midrule
  \multicolumn{2}{@{}l}{\textit{Layout Channel --- 15 features}}\\
  \midrule
  \texttt{found\_on\_page} & \textit{grounding.} $1$ if the value is grounded on the page ($|C_f|>0$), else $0$.\\
\texttt{cf\_count}       & \textit{grounding.} Number of candidate occurrences $|C_f|$.\\
\texttt{key\_found}      & \textit{grounding.} $1$ if a label anchor sits beside the value.\\
\texttt{match\_quality}  & \textit{grounding.} Token coverage of the value by its OCR span (strings). $1$ for exact-canon id/date/number.\\
\texttt{s\_l\_marg}      & \textit{spatial.} Marginal (log-mean-exp over occurrences) spatial log-likelihood --- the core score $s_L$.\\
\texttt{s\_l\_cold}      & \textit{spatial.} Cold-start spatial score: global field prior with retrieval off.\\
\texttt{s\_l\_abs}       & \textit{spatial.} Spatial score from absolute page position (weak baseline).\\
\texttt{anchor\_dist}    & \textit{geometry.} Distance to the nearest label anchor $z$.\\
\texttt{read\_rank}      & \textit{geometry.} Vertical reading-order position of the value.\\
\texttt{s\_match}        & \textit{retrieval.} Top retrieval similarity to a historical layout (template familiarity).\\
\texttt{sim\_margin}     & \textit{retrieval.} Top-1 minus top-2 retrieval similarity.\\
\texttt{k\_eff}          & \textit{retrieval.} Soft effective neighbour count from the retrieval weights.\\
\texttt{n\_eff}          & \textit{retrieval.} Effective neighbours contributing to the spatial prior.\\
\texttt{H\_f}            & \textit{ambiguity.} Entropy of the occurrence distribution $\pi_c$ (slot ambiguity).\\
\texttt{margin}          & \textit{ambiguity.} Top-two margin of $\pi_c$ (large = unambiguous).\\
\midrule
\multicolumn{2}{@{}l}{\textit{Validation Channel --- 8 features}}\\
\midrule
\texttt{v\_type\_ok}     & \textit{rule.} $1$ if the value has valid type/format for the field.\\
\texttt{v\_range\_ok}    & \textit{rule.} $1$ if the value lies in a plausible range.\\
\texttt{v\_soft}         & \textit{rule.} Soft aggregate $\texttt{v\_type\_ok}\times\texttt{v\_range\_ok}$ (always defined).\\
\texttt{v\_checksum}     & \textit{rule.} Checksum result where the field admits one (IBAN, ISIN); NaN otherwise.\\
\texttt{v\_arith}        & \textit{rule.} Arithmetic consistency (net $+$ tax $=$ gross; items vs.\ total); NaN if inapplicable.\\
\texttt{v\_applicable}   & \textit{rule.} $1$ if any hard validator (checksum or arithmetic) applies.\\
\texttt{v\_hard\_pass}   & \textit{rule.} $1$ if all applicable hard validators pass; NaN if none apply.\\
\texttt{xagree}          & \textit{cross-model.} $1$ if a second, independent VLM returns the same canonical value, else $0$.\\
\bottomrule
\end{tabular}
\caption{Feature dictionary for the fusion model, grouped by channel. Each entry leads with its sub-type (in italics). \texttt{NaN} rule features indicate the validator does not apply to that field (e.g.\ no check-summable identifier on an invoice).}
\label{tab:features}
\end{table*}

The canonicalizer $\mathrm{canon}_f$ defines both candidate matching and the ground-truth correctness label in section \ref{sec:statement}. An extraction $\hat{v}$ is correct iff it matches the gold value $v^\star$ under the per-category comparison below. We fix the ruleset at version \texttt{canon-v2} and treat it as a versioned artifact. 

\paragraph{Field categorization.}
Each field type is routed to one category below by substring rules on its name, tested in this order: date -> number -> identification. All others default to string.

\paragraph{Dates.}
We avoid full date parsing. We extract every run of digits and form the set of numeric components. A two-digit value $v \le 99$ contributes both itself and its century expansion ($2000{+}v$ if $v<50$, else $1900{+}v$). Month abbreviations (\texttt{JAN}-\texttt{DEC}) contribute their month index. Two dates match if their component sets overlap \emph{and} either one set is a subset of the other or they share at least two components. This order-agnostic rule is robust across the many formats in the data such as \texttt{3/4/2020}, \texttt{06/07/99}, \texttt{DEC31/01}, \texttt{JUL20/02}.

\paragraph{Numbers.}
We strip percent signs and any character outside $\{\texttt{0--9}, \texttt{.}, \texttt{,}, \texttt{-}\}$. The decimal separator is taken to be the rightmost \texttt{.} or \texttt{,} followed by one or two digits, all other separators are removed. The remainder is parsed as a signed float. Two numbers match under a strict \emph{absolute} tolerance of half a cent, $|a-b| \le 0.005$.

\paragraph{Identifiers and Strings.}
We strip accents (Unicode NFKD), lowercase, and remove all non-alphanumeric
characters. Two identifiers match on exact equality of this normalized form.

\begin{table*}[t]
\centering
\small
\setlength{\tabcolsep}{6pt}
\begin{tabular}{@{}lrrrlrrr@{}}
\toprule
Dataset & Indexed Layouts & Eval Doc & Field Types & VLM & Fields & Correct & Grounded\\
\midrule
\multirow{2}{*}{DocILE} & \multirow{2}{*}{5{,}180} & \multirow{2}{*}{500} & \multirow{2}{*}{19}
  & Qwen3.6-27B   & 5{,}141 & 51.7\% & 85.4\%\\
  & & & & Gemini-3.1-Flash-Lite & 5{,}077 & 53.8\% & 86.8\%\\
\midrule
\multirow{2}{*}{FATURA} & \multirow{2}{*}{731} & \multirow{2}{*}{200} & \multirow{2}{*}{14}
  & Qwen3.6-27B   & 1{,}850 & 60.2\% & 95.0\%\\
  & & & & Gemini-3.1-Flash-Lite & 1{,}767 & 63.9\% & 98.4\%\\
\midrule
\multirow{2}{*}{VRDU} & \multirow{2}{*}{417} & \multirow{2}{*}{200} & \multirow{2}{*}{9}
  & Qwen3.6-27B   & 1{,}698 & 63.3\% & 97.9\%\\
  & & & & Gemini-3.1-Flash-Lite & 1{,}666 & 63.7\% & 98.0\%\\
\bottomrule
\end{tabular}
\caption{Dataset statistics. \emph{Indexed Layouts} shows the unique number of training documents in the offline index. \emph{Eval Doc} is the number of evaluation documents. \emph{Types} is the number of distinct field types. \emph{Fields} is the number of labeled field extractions. \emph{Correct} is the fraction whose canonical value matches gold. \emph{Grounded} is the fraction recovered on the page ($|C_f|>0$). Note that the two VLMs extract slightly different fields.}
\label{tab:datastats}
\end{table*}

\section{Feature Dictionary}
\label{app:features}

The fusion model consumes 34 features across the three channels. Table~\ref{tab:features} lists all the features and the meanings. Features are computed at the best-scoring occurrence (Section~\ref{sec:statement}). When the value is not grounded ($|C_f|{=}0$), layout and pixel features fall back to fixed defaults and only the value-string and self-report features remain informative.

\section{Dataset Statistics}
\label{app:datastats}

Table~\ref{tab:datastats} reports the statistics for each dataset and VLM. The \emph{offline layout index} is built once per dataset from its training split (\emph{Indexed Layouts}), and is disjoint from the \emph{evaluation} documents (\emph{Eval}). The reporting unit is a field extraction (one row per (document, field)). Row counts differ slightly between the two VLMs on the same documents because grounding recovery depends on what each model extracted. DocILE uses its official validation split with template-familiarity buckets from train-cluster recurrence. FATURA and VRDU use 200-document evaluation samples. 

The three datasets are evaluated under a shared field taxonomy, with each dataset contributing the subset of field types it annotates unambiguously. 

\textbf{DocILE} \cite{vsimsa2023docile} uses the full set of 19 single-value field types, including 

- identifiers (document\_id, order\_id, customer\_id)

- dates (date\_issue, date\_due)

- monetary amounts (amount\_total\_gross,  amount\_total\_net, amount\_due, currency\_code\_amount\_due, amount\_paid, amount\_total\_tax)

- vendor fields (vendor\_name, vendor\_address, vendor\_tax\_id, vendor\_email)

- customer fields (customer\_billing\_name, customer\_billing\_address) 

- payment fields (payment\_terms, payment\_reference).

\textbf{FATURA} \cite{limam2023fatura} contains synthetic invoices annotating 24 classes. We map them into 14 with an unambiguous single-value analogue onto the shared taxonomy, by collapsing rate-specific tax classes and duplicate invoice-number classes. Layout regions (e.g., "table", "logo") and free-text blocks (e.g., "note", "payment\_details") have no clean field analogue and thus are excluded.

\textbf{VRDU} \cite{wang2023vrdu} contains real advertising order forms from a different domain. So its 9 fields are mapped by role onto the shared taxonomy. In particular, the "contract number" to "document\_id", the "product/estimate number" to "order\_id", the "campaign flight start and end" to "date\_issue" and "date\_due", the "gross total" to "amount\_total\_gross", the "TV station" to "vendor\_name/address", the advertiser to "customer\_billing\_name", and the "agency" to "customer\_other\_name".

\section{Layout Descriptor}
\label{app:descriptor}

The layout descriptor $\phi$ encodes a page's coarse text-occupancy geometry, independent of the text content. We overlay a $G\times G$ grid ($G{=}32$) on the page in normalized coordinates and mark every cell that any OCR token box overlaps, giving a binary occupancy mask; the mask is flattened to a $G^2{=}1024$-dimensional vector and L2-normalized. Two pages are compared by the inner product of their descriptors, which equals cosine similarity after normalization, so the FAISS inner-product index returns the layouts whose text falls in the same places. One descriptor is computed per document from all its OCR words (across pages).

The descriptor is deliberately coarse and content-blind. It fingerprints \emph{where} fields sit on the page, not what they say. This is exactly the
invariance retrieval needs. Two invoices from the same vendor template map to nearby descriptors regardless of their amounts, dates, or names, so the spatial prior for a field can borrow strength across all documents that share its layout. A cell is set whenever a token box overlaps it, with an inclusive end cell, so a text line thinner than one grid cell still marks its cell rather than being dropped by an exclusive slice. Listing~\ref{lst:descriptor} gives the details of the implementation.

\begin{lstlisting}[style=pycode,label={lst:descriptor},caption={Layout descriptor $\phi$ is a binary $G\times G$ OCR-token occupancy mask, flattened and L2-normalized. Word boxes are in normalized page coordinates.}]
def token_mask_descriptor(words, grid=32):
    """Binary grid x grid OCR-token occupancy mask, L2-normalized."""
    m = np.zeros((grid, grid), np.float32)
    for w in words:
        l, t, r, b = w.bbox 
        l = min(max(l, 0.0), 1.0)
        r = min(max(r, 0.0), 1.0)
        t = min(max(t, 0.0), 1.0)
        b = min(max(b, 0.0), 1.0)
        x0 = min(int(l * grid), grid - 1)
        x1 = min(int(r * grid), grid - 1)
        y0 = min(int(t * grid), grid - 1)
        y1 = min(int(b * grid), grid - 1)
        # cells the token overlaps
        m[y0:y1 + 1, x0:x1 + 1] = 1.0   
        
    # flatten to grid * grid
    v = m.reshape(-1)                         
    n = np.linalg.norm(v)
    if n > 0:
        v = v / n                 
    return v.astype(np.float32)
\end{lstlisting}

At query time, a descriptor is issued against the index to retrieve the $K_{\mathrm{NN}}{=}50$ most similar layouts. Their similarities become the retrieval weights that condition the spatial prior. The same encoder is used offline to build the index and online to form the query, so familiarity is measured consistently.

\section{VLM Extraction Prompt}
\label{app:prompt}

Each document is extracted in a single user message: the page image followed by the following instruction. The model returns strict JSON with a value and verbalized confidence per present field, omitting absent ones. The same template is used for both VLMs. Only a per-family ``thinking'' toggle differs (Qwen has \texttt{enable\_thinking=false} and Gemini has \texttt{reasoning\_effort=none}). The fields listed below are the DocILE schema. FATURA and VRDU use the same template with their own mapped fields. The amount rules are appended to resolve the common net/gross/due confusion.

\begin{lstlisting}[style=prompt,label={lst:prompt},caption={The VLM extraction prompt with DocILE field schema as the example.}]
Extract the following fields from this business document image.
Return STRICT JSON only, no prose. For each field that is PRESENT on the page, give an object {"value": <exact text as printed>, "confidence": <integer 0-100>}. OMIT fields that are absent.
Fields:
  - document_id: the invoice/document number
  - order_id: the purchase order number
  - date_issue: the issue/invoice date
  - date_due: the payment due date
  - amount_total_gross: grand total INCLUDING tax (equals net + tax)
  - amount_total_net: subtotal BEFORE tax (net of tax)
  - amount_total_tax: the TAX/VAT amount only (not a total)
  - amount_due: amount actually payable now (often equals gross)
  - amount_paid: amount already paid
  - currency_code_amount_due: currency symbol or ISO code
  - vendor_name: the seller/vendor company name
  - vendor_address: the full vendor postal address
  - vendor_tax_id: vendor tax / VAT id
  - vendor_email: vendor email address
  - customer_billing_name: the buyer/customer billing name
  - customer_billing_address: the full customer billing address
  - customer_id: the customer account id
  - payment_terms: payment terms (e.g. NET 30)
  - payment_reference: payment reference / variable symbol
Respond as: {"document_id": {"value": "...", "confidence": 90}, ...}

Amount rules: net = subtotal before tax; tax = the tax amount only; gross = grand total including tax (gross = net + tax); due = amount payable now. If the page shows a single total, use it for BOTH gross and due. Report each amount exactly as printed.
\end{lstlisting}

\paragraph{Sampling.}
Each document is queried $K_q{=}5$ times. The first sample is greedy
(temperature $0$) and fixes the reported value, the remaining four are sampled (temperature $0.7$) to estimate self-consistency. The reported value is the modal canonical value across samples, the verbalized confidence $s_f$ is the mean self-reported confidence, and the sampling-agreement fraction is the self-consistency baseline. The deployed operating point is the single greedy call. The additional samples are used only to compute the self-consistency baseline.

\section{Fusion model configuration}
\label{app:fusion}

\begin{table}[t]
\centering
\small
\setlength{\tabcolsep}{6pt}
\begin{tabular}{lll}
\toprule
Hyperparameter & Value \\
\midrule
objective            & binary log-loss \\
boosting             & \texttt{gbdt}   \\
\texttt{n\_estimators}                  & 200    \\
\texttt{learning\_rate}                 & 0.08   \\
\texttt{num\_leaves}                    & 15    \\
\texttt{reg\_lambda L2 regularization}  & 1.0   \\
\texttt{min\_child\_samples}            & 20    \\
\texttt{random\_state}                  & 0      \\
\bottomrule
\end{tabular}
\caption{LightGBM fusion-model configuration.}
\label{tab:lgbm}
\end{table}

The fusion model is a gradient-boosted decision tree (\texttt{LightGBM.LGBMClassifier}) with the binary log-loss objective. Table~\ref{tab:lgbm} lists the full configuration. Regularization is deliberately conservative for the small tabular problem ($\sim$800--2600 training rows per split), shallow trees ($15$ leaves), an L2 penalty, a minimum of $20$ samples per leaf and no row or feature subsampling. The model is retrained from scratch on each document-level split.

\section{Qualitative Case Studies}
\label{app:cases}

\begin{figure*}[!tb]
\centering
\begin{subfigure}[t]{0.49\textwidth}
  \includegraphics[width=\textwidth]{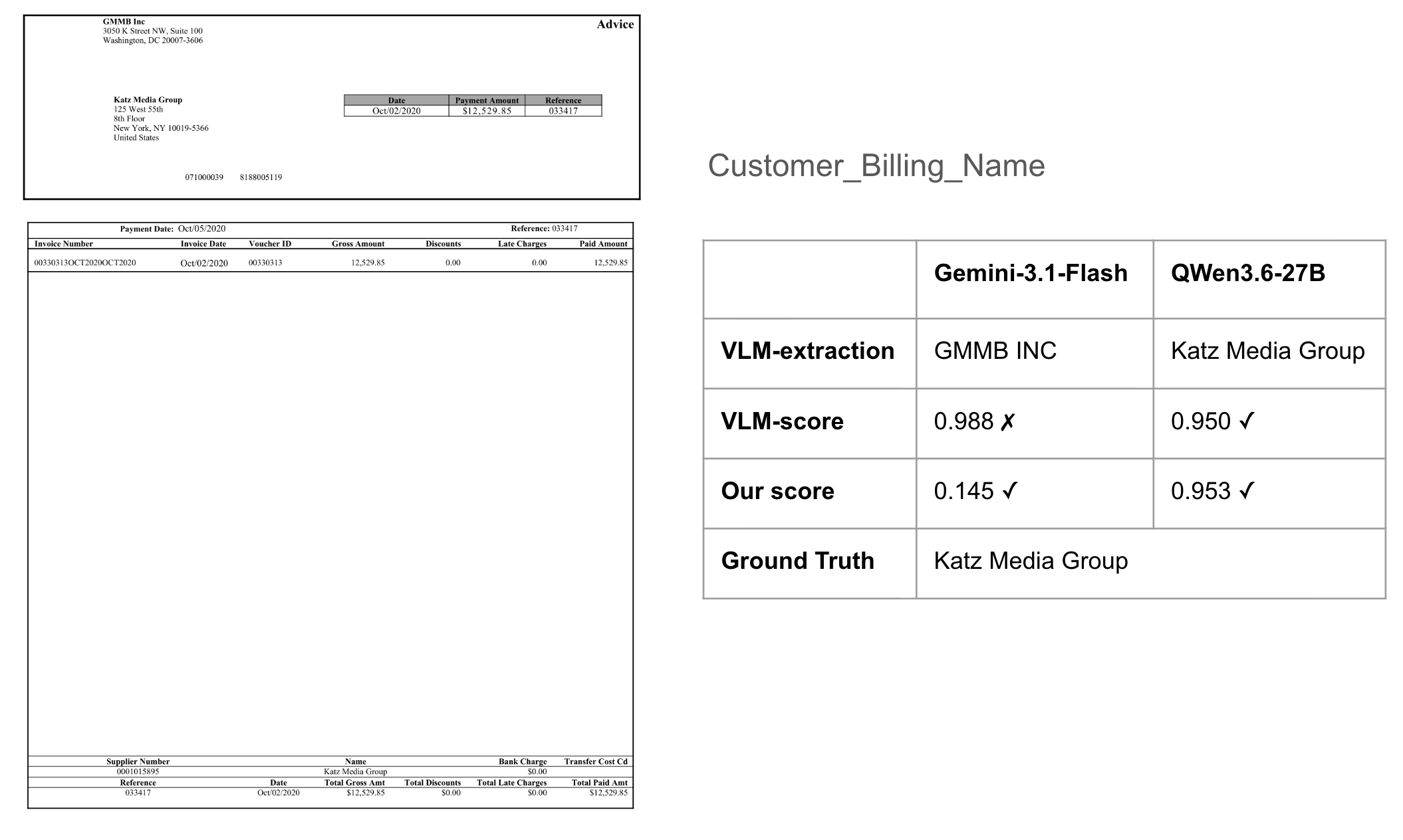}
  \caption{Sender vs. recipient confusion (\texttt{customer\_billing\_name}): fused
  $0.145$ vs.\ verbalized $0.953$ on Gemini result.}
  \label{fig:case1}
\end{subfigure}\hfill
\begin{subfigure}[t]{0.49\textwidth}
  \includegraphics[width=\textwidth]{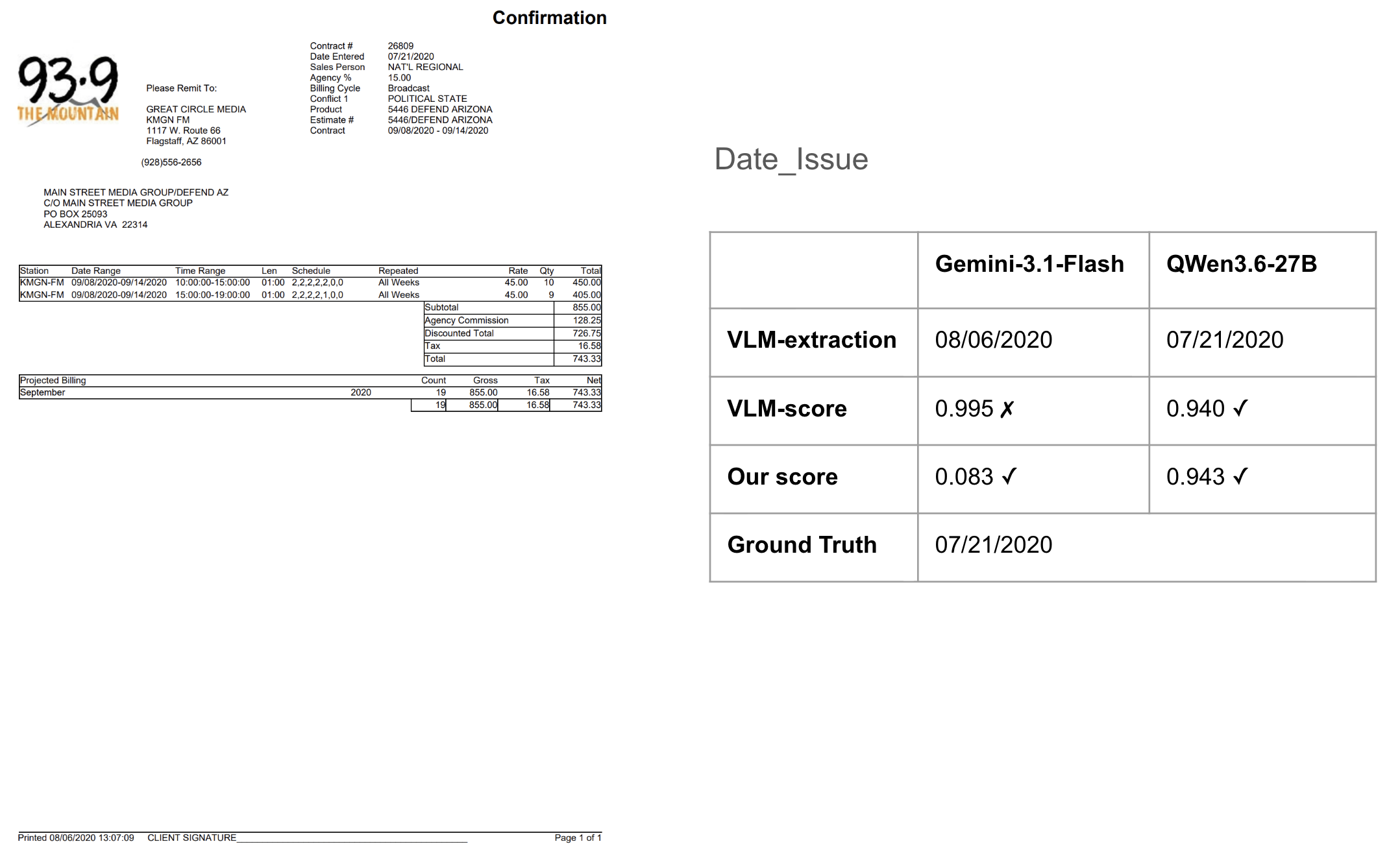}
  \caption{Print-timestamp vs.\ issue date (\texttt{date\_issue}): fused $0.083$ vs.\ verbalized $0.943$ on Gemini result.}
  \label{fig:case2}
\end{subfigure}

\vspace{0.6em}

\begin{subfigure}[t]{0.49\textwidth}
  \includegraphics[width=\textwidth]{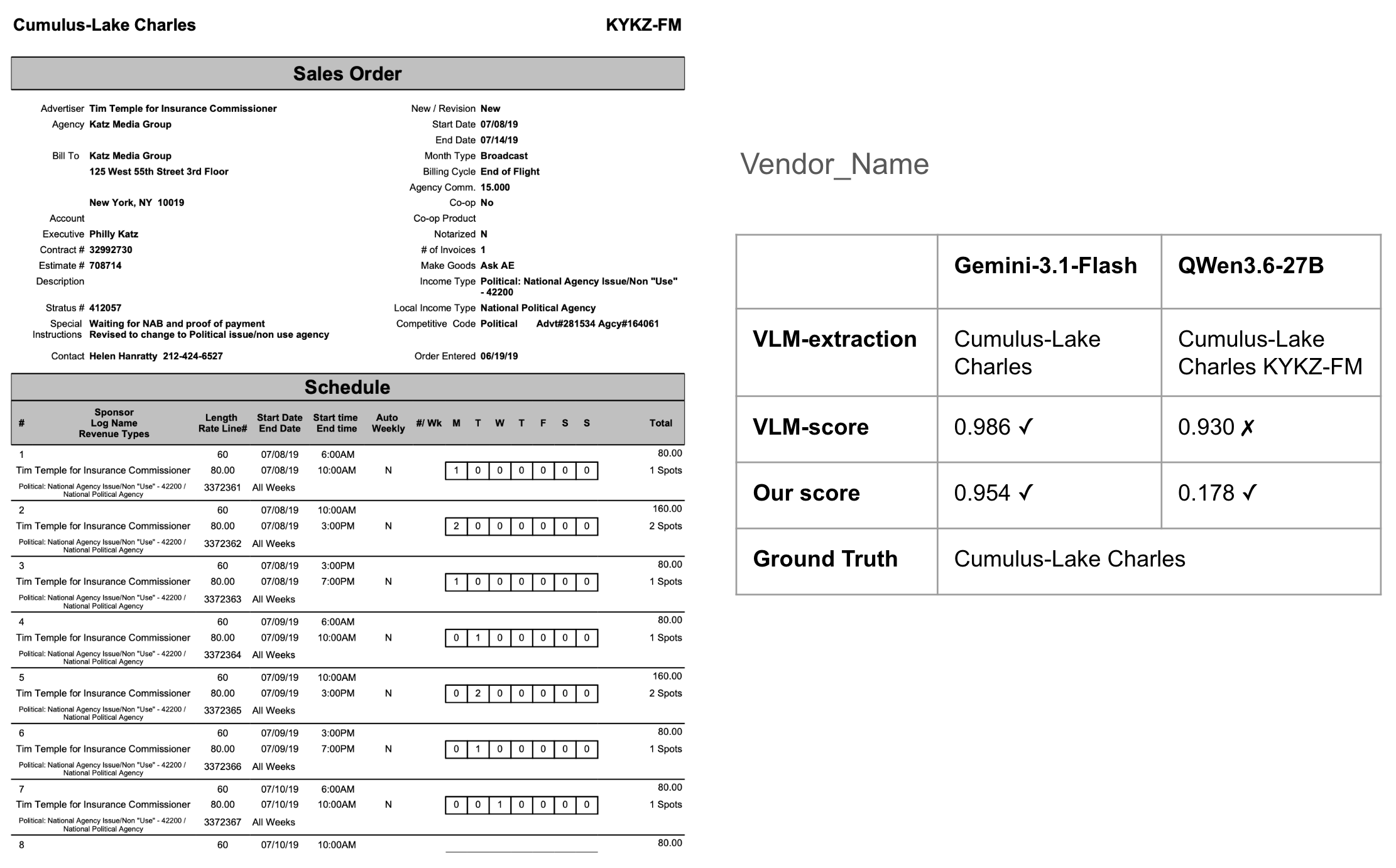}
  \caption{Call-sign over-inclusion (\texttt{vendor\_name}): fused $0.178$ vs.\ verbalized $0.954$ on Qwen result.}
  \label{fig:case3}
\end{subfigure}\hfill
\begin{subfigure}[t]{0.49\textwidth}
  \includegraphics[width=\textwidth]{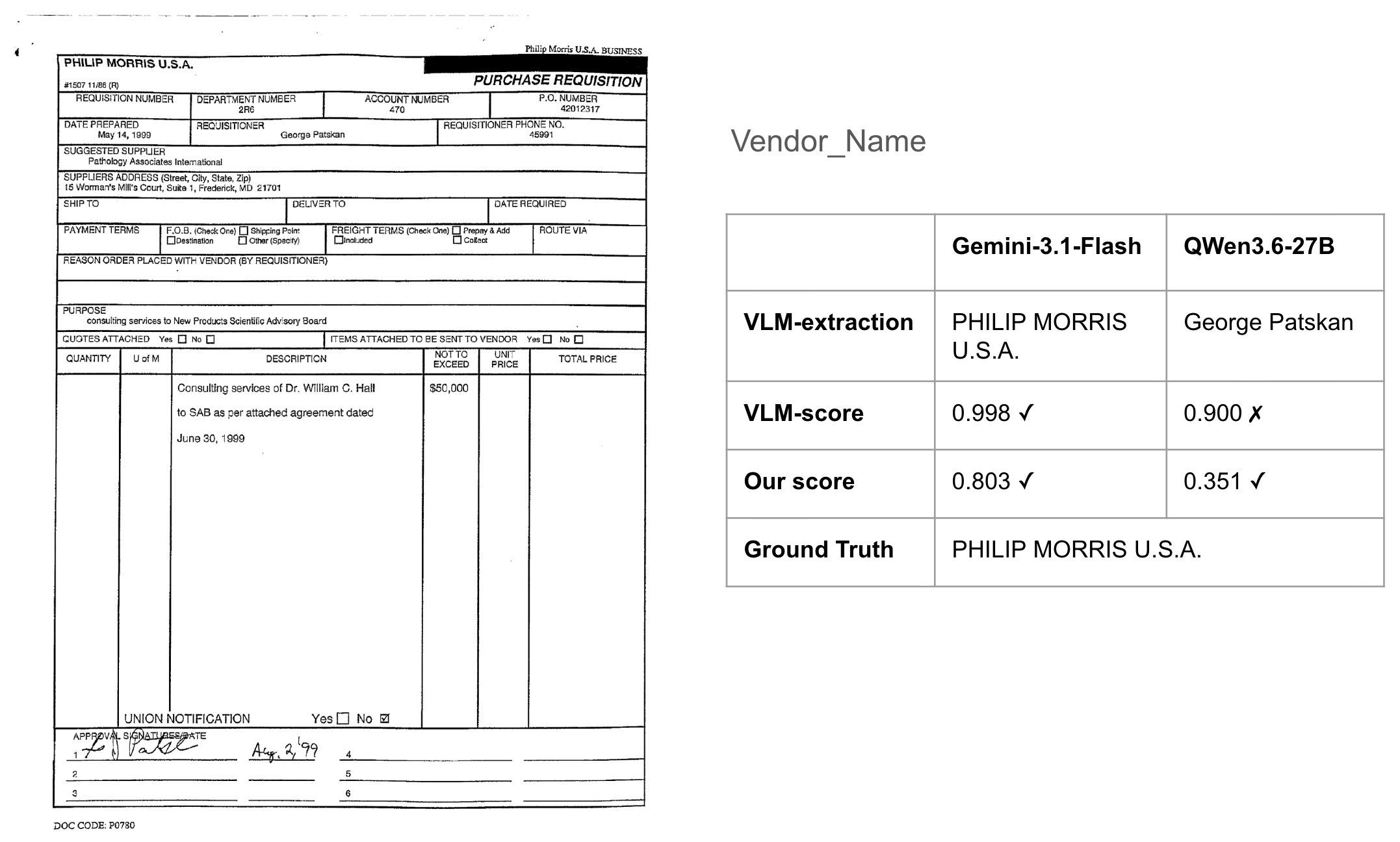}
  \caption{Person vs.\ company (\texttt{vendor\_name}): fused $0.351$ vs.\ verbalized $0.803$ on Qwen result.}
  \label{fig:case4}
\end{subfigure}
\caption{Qualitative case studies. In each, one VLM is confidently wrong (verbalized $\ge 0.90$) and the other is correct. Our fused score separates them (wrong value listed first). The confidently-wrong extraction is assigned a low fused probability in every case, while the correct value scores high.}
\label{fig:cases}
\end{figure*}

We show four cases where one VLM extracts a wrong value with high verbalized confidence, but the confidence layer assigns it a low probability. In every case the wrong and right values have nearly indistinguishable verbalized confidence ($0.90$--$0.998$), yet the fused score separates them cleanly. This is the qualitative counterpart of the aggregate AUROC gap. It also shows why cross-model selection works by keeping the value our layer scores higher to recover the correct answer in all four cases. Each case illustrates a different error mode.

\paragraph{Case 1. Sender vs. recipient confusion in Figure~\ref{fig:case1} on \texttt{customer\_billing\_name} field.} On a payment-advice document, Gemini returns the paying party "GMMB Inc" (verbalized $0.988$) instead of the billed customer "Katz Media Group" (gold). Qwen returns the correct value ($0.950$). Verbalized confidence cannot tell them apart, but our layer scores Gemini's value $0.145$ and Qwen's $0.953$: the sender's name sits in the letterhead block, where a billing-name field rarely lands.

\paragraph{Case 2. Wrong date field as shown in Figure ~\ref{fig:case2} on \texttt{date\_issue} extraction.} On a broadcast confirmation, Gemini returns "08/06/2020" (verbalized $0.995$), which is the footer \emph{print} timestamp, not the "Date Entered" issue date "07/21/2020" (gold). The reading is crisp and unambiguous, so perception signals do not help; the layout channel penalizes the value because a footer position is far from where an issue date normally appears, scoring it $0.083$ against Qwen's $0.943$.

\paragraph{Case 3. Call-sign over-inclusion in Figure~\ref{fig:case3} on \texttt{vendor\_name}.} Qwen returns "Cumulus-Lake Charles KYKZ-FM" concatenating the station call sign onto the vendor name (verbalized $0.930$), while gold is "Cumulus-Lake Charles" aligned by Gemini. Our layer scores Qwen's over-inclusive value $0.178$ and Gemini's value $0.954$, driven by the span/OCR-coverage and value-string cues.

\paragraph{Case 4. Person vs. company in Figure \ref{fig:case4} on \texttt{vendor\_name}.} On a purchase requisition, Qwen returns the requisitioner person name "George Patskan" (verbalized $0.900$) instead of the vendor company "PHILIP MORRIS U.S.A." (gold). Our layer scores Qwen's $0.351$ and Gemini's $0.803$, where the requisitioner name sits in an internal form field, not where a vendor name is expected.